%% file: main.tex
\documentclass[letterpaper]{article} % DO NOT CHANGE THIS
\usepackage{aaai25}  % DO NOT CHANGE THIS
\usepackage{times}  % DO NOT CHANGE THIS
\usepackage{helvet}  % DO NOT CHANGE THIS
\usepackage{courier}  % DO NOT CHANGE THIS
\usepackage[hyphens]{url}  % DO NOT CHANGE THIS
\usepackage{graphicx} % DO NOT CHANGE THIS
\usepackage{natbib}  % DO NOT CHANGE THIS AND DO NOT ADD ANY OPTIONS TO IT
\usepackage{caption} % DO NOT CHANGE THIS AND DO NOT ADD ANY OPTIONS TO IT
\usepackage{algorithm}
\usepackage{algorithmic}

\usepackage{booktabs} % For formal tables
\usepackage{graphicx}
\usepackage{subfigure}
\usepackage{epstopdf}
\usepackage{epsfig}
\usepackage{xspace}
\usepackage{amssymb,amsmath}
\usepackage{enumitem}
\usepackage{algorithm}
\usepackage{algorithmic}

\usepackage{amsthm}
\usepackage{caption}
\usepackage{multirow}

\usepackage{threeparttable}
\usepackage{diagbox}

\usepackage{url}

\newtheorem{Def}{Definition}

\newtheorem*{Pro*}{Problem}
\usepackage{newfloat}
\usepackage{listings}
\DeclareCaptionStyle{ruled}{labelfont=normalfont,labelsep=colon,strut=off} % DO NOT CHANGE THIS
\floatstyle{ruled}
\newfloat{listing}{tb}{lst}{}
\floatname{listing}{Listing}
\title{Scalable Trajectory-User Linking with Dual-Stream Representation Networks}
\author{
    Hao Zhang\textsuperscript{\rm 1}\equalcontrib,
    Wei Chen\textsuperscript{\rm 2}\equalcontrib,
    Xingyu Zhao\textsuperscript{\rm 1},
    Jianpeng Qi\textsuperscript{\rm 1},
    Guiyuan Jiang\textsuperscript{\rm 1},
    Yanwei Yu\textsuperscript{\rm 1}\thanks{Corresponding author: Yanwei Yu.}
}
\affiliations{
    \textsuperscript{\rm 1}Faculty of Information Science and Engineering, Ocean University of China, Qingdao, China\\
    \textsuperscript{\rm 2}The Hong Kong University of Science and Technology (Guangzhou)
    \{zhanghao9222,zhaoxingyu\}@stu.ouc.edu.cn,onedeanxxx@gmail.com,\\
    \{qijianpeng,jiangguiyuan,yuyanwei\}@ouc.edu.cn

}

\usepackage{bibentry}
\begin{document}

\maketitle
\newcommand{\model}{ScaleTUL\xspace}
\begin{abstract}

Trajectory-user linking (TUL) aims to match anonymous trajectories to the most likely users who generated them, offering benefits for a wide range of real-world spatio-temporal applications. However, existing TUL methods are limited by high model complexity and poor learning of the effective representations of trajectories, rendering them ineffective in handling large-scale user trajectory data.
In this work, we propose a novel \underline{Scal}abl\underline{e} Trajectory-User Linking with dual-stream representation networks for large-scale \underline{TUL} problem, named \model. Specifically, \model generates two views using temporal and spatial augmentations to exploit supervised contrastive learning framework to effectively capture the irregularities of trajectories. In each view, a dual-stream trajectory encoder, consisting of a long-term encoder and a short-term encoder, is designed to learn unified trajectory representations that fuse different temporal-spatial dependencies. Then, a TUL layer is used to associate the trajectories with the corresponding users in the representation space using a two-stage training model. 
Experimental results on check-in mobility datasets from three real-world cities and the nationwide U.S. demonstrate the superiority of \model over state-of-the-art baselines for large-scale TUL tasks. 

\end{abstract}


\input{1Introduction}
\input{2RelatedWork}

\input{3Preliminaries}

\input{4methodology}
\input{5Experiments}
\input{6Conclusion}

% \section{Acknowledgments}
\section*{Acknowledgments}
This work is partially supported by the National Natural Science Foundation of China under Grant Nos. 62176243 and 62372421, the Fundamental Research Funds for the Central Universities under Grant No 202442005,  the National Key R\&D Program of China under Grant No 2022ZD0117201, and Shandong Postdoctoral Innovative Talent Support Program under Grant No. SDBX2023013.

% \bigskip
% \noindent Thank you for reading these instructions carefully. We look forward to receiving your electronic files!

% \bibliographystyle{aaai25}
\bibliography{main}

\end{document}

%% file: 1Introduction.tex
\section{Introduction}

The proliferation of various location-based service platforms has generated a vast amount of user mobility data, offering researchers valuable opportunities to understand and analyze human movement patterns~\cite{wang2020deep}. Trajectory-user linking (TUL) aims to associate given anonymous trajectories with the most likely users who generated them~\cite{gao2017identifying}. This problem has garnered significant research interest in recent years because effective TUL methods can benefit a wide range of spatio-temporal applications, including personalized location recommendations~\cite{liu2019geo}, route suggestions~\cite{7733273}, behavior tracking~\cite{chen2019human}, and the identification of terrorist or criminal activities~\cite{huang2018deepcrime}. Furthermore, such methods can also be used to test and challenge privacy protection mechanisms in mobility data analysis models~\cite{rao2020lstm}.

In this paper, we focus on addressing the large-scale TUL problem using deep learning models. Traditional methods rely mainly on similarity measurement techniques, such as LCSS~\cite{ying2010mining}, STS~\cite{li2021spatial}, and SR~\cite{jin2019moving,jin2020trajectory}, to assess the similarity between trajectories and link anonymous trajectories with users of similar ones. Early deep learning models typically employ deep neural networks, such as Seq2Seq~\cite{gao2017identifying}, Variational Autoencoders (VAE)~\cite{zhou2018trajectory}, and Recurrent Neural Networks (RNNs)~\cite{medsker2001recurrent}, to learn spatiotemporal representations of trajectories and model the problem as a trajectory classification task. Recently, models like MainTUL~\cite{chen2022mutual} and S$^2$TUL~\cite{deng2023s2tul} have been proposed, utilizing mutual distillation learning networks and graph neural networks to capture spatiotemporal movement patterns and relationships between trajectories for TUL problem, respectively. Additionally, a greedy trajectory-user re-linking method has been designed to address the issue of time conflicts in S$^2$TUL. 
 
Despite the notable progress achieved by these methods, they still face two major limitations. 
\textit{First}, existing methods mainly focus on the TUL problem on hundreds of users using trajectory classification methods, while real-world application scenarios are large-scale and involve massive trajectory users. Current deep leaning models are not specifically designed for large-scale user trajectory data. They are typically formulated as multi-class trajectory classification problems, which results in inefficiencies and limited performance~\cite{joulin2017efficient}. 
\textit{Second}, most TUL methods directly learn the representation of trajectories divided by day or longer historical trajectories, lacking effective exploration of the irregularity of trajectories and failing to effectively capture various different spatio-temporal dependencies in large-scale user trajectories. 

To address these challenges, we propose \model, a \underline{Scal}abl\underline{e} Trajectory-User Linking with dual-stream representation networks for large-scale \underline{TUL} problem. 
Specifically, we design a spatio-temporal augmentation strategy that includes both temporal and spatial augmentations. The temporal augmentation introduces trajectories with different time granularities to fully capture user movement patterns across various periods, while the spatial augmentation addresses trajectory sparsity and enhances the potential irregularities of trajectories to improve model robustness. Additionally, we develop a dual-stream trajectory encoder, consisting of a structured stae space model~\cite{gu2023mamba} to capture long-term spatio-temporal dependencies in temporal extended trajectories and a recurrent neural network variant~\cite{graves2012long} to learn short-term spatio-temporal movement patterns. Through dual-view contrastive learning, the model effectively learns the aligned and consistent trajectory representations and the corresponding user label representations. Finally, a two-stage training process is employed to link the trajectories with their corresponding user labels in the representation space. 

On three cities and across the nationwide United States check-in dataset from Foursquare, our experiments show that \model significantly outperforms state-of-the-art baselines in TUL task. 

Our contributions can be summarized as follows:
\begin{itemize}
    \item We propose a novel scalable trajectory-user linking with dual-stream representation network, named \model. Our \model effectively and efficiently addresses the challenge of large-scale TUL.
 
    \item We design a dual-stream trajectory encoder that includes two distinct encoders, effectively capturing both the long-term and short-term spatio-temporal dependencies of user trajectories, in combination with the spatiotemporal augmentation strategy. 
    \item 
    We conduct extensive experiments on check-in mobility data from three cities as well as across the entire United States. The experimental results demonstrate the effectiveness and superiority of our model in addressing large-scale TUL problem against state-of-the-art baselines.

\end{itemize}

%% file: 2RelatedWork.tex
\section{Related Work}
To address the TUL problem, researchers have developed various approaches, which can be broadly categorized into three groups:

\textbf{Similarity measure-based methods.} Traditionally, the TUL problem is addressed through trajectory similarity retrieval methods~\cite{magdy2015review,sousa2020vehicle}. 

Commonly used similarity measures include Longest Common Subsequence (LCSS) ~\cite{ying2010mining}, dynamic time warping (DTW)~\cite{keogh2000scaling}, Signature Representation (SR)~\cite{jin2019moving}, and spatial-temporal similarity (STS)~\cite{li2021spatial}. \textit{However, due to the complexity of similarity computations, these approaches tend to be computationally expensive and time-consuming.}

\textbf{Statistical learning-based methods.} Typically, these methods extract statistical features from trajectory data and train supervised models to predict the likelihood that a user generated a given trajectory ~\cite{pao2012trajectory,ren2020st}. For instance, ~\cite{ren2020st} employs both profile features and online features derived from trajectory data to train a binary classifier for determining if two trajectories belong to the same user. \cite{najjar2022trajectory} is the first work to address the challenge of large-scale TUL by using a non-parametric classifier. \textit{Such models' effectiveness depends on hand-crafted features' quality, which relies on domain knowledge and understanding of the trajectory dataset.}

\textbf{Deep learning-based methods}.
Thanks to the powerful learning capabilities of neural networks, numerous deep learning-based methods have been developed to address the TUL problem using check-in trajectory data~\cite{10.1145/3635718,chen2024deep}. These approaches primarily focus on learning high-order representations of user mobility patterns. For instance, TULER~\cite{gao2017identifying} employs RNNs like LSTM to learn sequential transition patterns, while TUAVE~\cite{zhou2018trajectory} uses a VAE with RNNs to capture hierarchical semantics. DeepTUL~\cite{miao2020trajectory} leverages a bidirectional LSTM with attention for learning periodic mobility patterns, and GAN-based models such as AdattTUL~\cite{gao2020adversarial}, TGAN~\cite{zhou2021improving}, and TULMAL~\cite{zhang2022multi} have also been proposed. More recently, MainTUL~\cite{chen2022mutual} utilizes knowledge distillation with RNN and Transformer encoders, S$^2$TUL~\cite{deng2023s2tul} constructs graphs to integrate trajectory-level information, and SAMLink~\cite{chen2023samlink} designs adaptive fusion modules for different types of trajectory data.
\textit{However, previous studies have been limited to classifying trajectories for at most a few hundred users, without addressing scalability problem. This limitation stems from the use of complex neural network architectures combined with classification layers, which are both time-inefficient and memory-intensive.}

%% file: 3Preliminaries.tex
\section{Preliminaries}
\begin{figure*}[t!]
    
    \begin{center}
    \includegraphics[width=0.96\textwidth]{  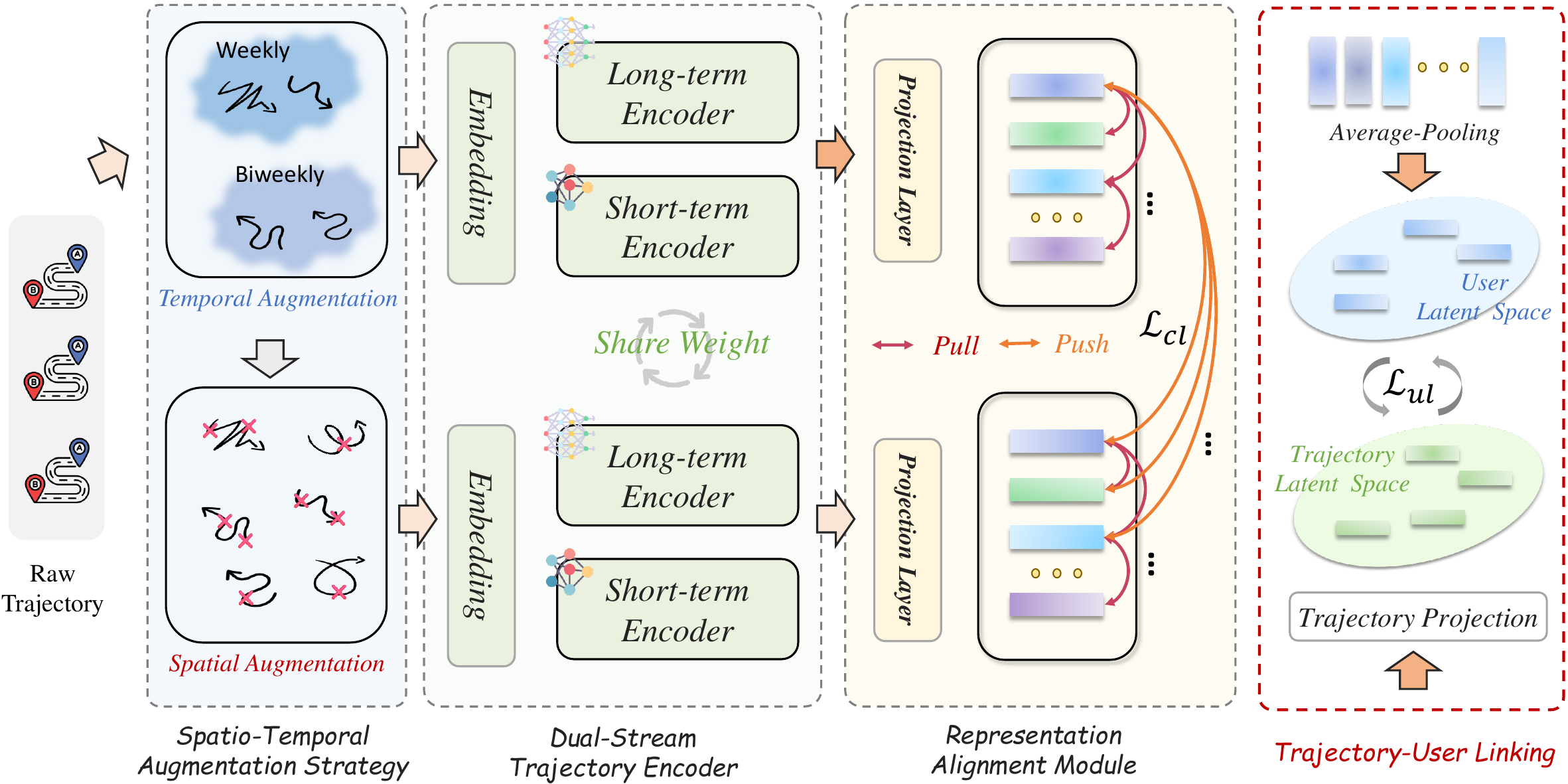}
    % \vspace{-1mm}
    \caption{The overview of the proposed framework.}
    \vspace{-3mm}
    \label{fig:overview}
    \end{center}
\end{figure*}

\begin{Def}(Check-in Trajectory).
    A check-in trajectory is a chronological sequence of check-in records, where each record is represented as a tuple $(u, p, c, t)$. In this tuple, $u$ represents the user, $p$ denotes a uniquely identified venue, $c$ is the category of the venue, and $t$ is the timestamp of the check-in. The trajectory, denoted as $Tr_{\tau}$, is generated by a user $u$ within a specific time interval $\tau$. 
\end{Def}
When the user associated with a trajectory is unknown, it is referred to as an unlinked trajectory. 
In this paper, we define the time interval $\tau$ as one day. The Trajectory-User Linking (TUL) problem aims to map these unlinked trajectories to their corresponding users.

\begin{Pro*}(Trajectory-User Linking). The TUL task aims to link anonymous trajectories to their respective users. Let $\mathcal{T} = \{Tr^1, Tr^2, \dots, Tr^n\}$ be the set of unlinked trajectories, and $\mathcal{U} = \{u_1, u_2, \dots, u_m\}$ represent the set of users.

Our goal is to identify a mapping function $f(\cdot)$ that links unknown trajectories to the corresponding users: $\mathcal{T} \mapsto \mathcal{U}$. 
This function should map each trajectory to a representation that reflects the user's identity, ensuring correct association with the rightful user.

\end{Pro*}

%% file: 4methodology.tex
\section{Methodology}

In this section, we present detailed introduction about the trajectory user linking framework based on dual-stream networks (as illustrated in Figure~\ref{fig:overview}). \model is accomplished through a two-stage process involving representation learning and user linking. We delve into how to learn high-quality trajectory representations from the perspectives of diverse feature capture and lightweight trajectory encoding. Specifically, we first introduce three key components of trajectory representation: \textit{Spatio-Temporal Augmentation Strategy}, \textit{Dual-Stream Trajectory Encoder}, and \textit{Representation Alignment Module}. Then, we introduce how to efficiently link representations in trajectory and user latent spaces.

\subsection{Spatio-Temporal Augmentation Strategy}

Raw trajectory data of human mobility is often sparse and unable to capture the diverse underlying movement intentions. To address this problem, we propose two augmentation strategies from both temporal and spatial perspectives.

\subsubsection{Temporal Augmentation.}
Human mobility information typically exhibits periodic characteristics over time. To fully capture users' temporal visitation preferences and movement patterns across different scales, we propose a periodic temporal augmentation strategy. Specifically, the raw check-in sequences are often segmented into daily trajectories, denoted as $\{{Tr}_{\tau}|\tau=1\}$ . We then query the trajectory sequences of the current user for the upcoming week and the next two weeks, denoted as $\{{Tr}_{\tau}|1 \leq \tau \leq 7\}$ and $\{{Tr}_{\tau}|1 \leq \tau \leq 14\}$, respectively. Subsequently, the temporal augmented trajectories from different periodic are consolidated into new training samples.

\subsubsection{Spatial Augmentation.}
Due to the varying check-in preferences of users across different locations and the potential for communication failures in recording devices, the resulting check-in sequences often exhibit low-sampled trajectories with significant missing locations. To effectively simulate the irregular mobility patterns and noise characteristics that lead to such sparsely sampled movement data, we propose a random spatial augmentation strategy. Specifically, for each newly generated check-in trajectory sample derived from temporal augmentation, we further apply a random location masking technique to produce a corresponding augmented trajectory, denoted as ${Tr}^{\prime}_{\tau}={Tr}_{\tau} \odot {M}_{r}$. Here, ${M}_{r}$ represents a binary mask vector, where the masking rate $r$ is defined as the ratio of the number of masked location points to the total length of the trajectory.

It is worth noting that these two augmentation strategies are applied only during the training phase. By employing these two mobility data augmentation strategies, not only is the issue of sparse training data alleviated from a spatial perspective, but the model is also better able to capture more diverse movement patterns from a temporal perspective.

\subsection{Dual-Stream Trajectory Encoder}

Due to the design of the augmentation strategies, significant differences in sequence lengths between different trajectories are inevitable, resulting in a multi-modal distribution. Consequently, a simple trajectory encoder is insufficient for modeling such overlapping distribution characteristics. To address this limitation, we propose a dual-stream trajectory encoder, which includes a unified embedding layer along with separate short-term and long-term encoders, each designed to model patterns with different length distributions.

\subsubsection{Unified Embedding Layer.}
Each check-in location of trajectory typically includes not only geographic information but also rich external information, such as time and semantic categories. Therefore, before capturing sequential patterns, it is necessary to integrate all the above information into a unified location embedding. Here, we use a simple nonlinear layer fusion to obtain the semantic location representation:
\begin{equation}
x_{i}= f(\theta;~p_i,~t_i,~c_i),
\end{equation}
where $p$, $t$, and $c$ denote the discrete mapping of location $p$ at time $t$ with category $c$. The function $f_{\theta}$ is a simple multi-layer perceptron with a nonlinear activation function.

The complete embedding matrix for a single trajectory can then be obtained by stacking the representations of all individual locations:
\begin{equation}
\mathbf{X}=\{x_{i}|i=1,2,\ldots,n\}.
\end{equation}

By doing so, each trajectory is preliminarily mapped into a high-dimensional semantic representation.

\subsubsection{Dual-Stream Encoder.}
As previously mentioned, the augmented trajectories typically exhibit significant differences in sequence lengths. Therefore, it is necessary to design appropriate trajectory encoders for these two different lengths of trajectories to capture their respective features. Additionally, due to the introduction of more trajectory encoders, to maintain efficiency in the sequence learning process, it is essential to select sufficiently lightweight encoding models to support the training of large-scale trajectory data.

To model short-term movement patterns, we select a simple recurrent neural network variant, Bi-directional Long Short-Term Memory~\cite{graves2012long}, as the short-term encoder $\mathcal{F}_{\theta_{\alpha}}$. Furthermore, we use the hidden layer output at the last time step to represent the dynamic pattern of the short trajectory:
\begin{equation}
z_{\alpha}= \mathcal{F}(\theta_{\alpha};~\mathbf{X}).
\end{equation}

Unlike the simple short-term patterns, long-term trajectory patterns require selective processing of input location information while also satisfying long-range dependency modeling. Recently, structured state space models~\cite{gu2023mamba} have gained attention due to their hardware-aware lightweight and efficient characteristics, which fulfill our requirements. Therefore, we select them as the long-term encoder $\mathcal{F}_{\theta_{\beta}}$. Furthermore, we use the hidden state at the last time step as the long trajectory representation:
\begin{equation}
z_{\beta}= \mathcal{F}(\theta_{\beta};~\mathbf{X}).
\end{equation}

Then we use a learnable parameter $\gamma$ to balance the weights of the representations from two encoders to obtain the final representation, and use ridge normalization to standardize the trajectory representation, improving numerical stability and accelerating the model’s training process:
\begin{equation}
{z}=\|(\gamma \cdot {z_{\alpha}}+(1-\gamma) \cdot {z_{\beta}})\|_2.
\end{equation}

\subsection{Representation Alignment Module}

Intuitively, by applying spatial augmentation strategies, we obtain two views of trajectory samples. A natural approach is to align the masked-view trajectory with the unmasked-view trajectory via contrastive learning to ensure robust trajectory representation~\cite{LI2023357}. However, unlike traditional self-supervised learning, which relies on contrastive learning with negative samples, our unique advantage lies in the accessibility of trajectory labels. Therefore, in addition to basic negative sample information, we can also consider the associated positive sample signals for each anchor (in contrast to traditional self-supervised contrastive learning that typically uses only a single positive sample signal). By leveraging labeled data to enhance contrastive learning, we can obtain higher-quality trajectory representations, thereby avoiding the misclassification of the same user's trajectories as negative samples.

Specifically, given a batch of $N$ unmasked random sample-label pairs, the corresponding $2N$ multi-view sample pairs are generated through spatial masking augmentation strategies. We then apply an extended supervised version~\cite{khosla2020supervised} of the traditional Noise Contrastive Estimation loss (InfoNCE)~\cite{chen2020simple} to achieve more effective representation alignment:
\begin{equation}
 \begin{split}
    \mathcal{L}_{cl} &= \sum_{i\in 2N}\frac{-1}{|P(i)|}\sum_{p\in P(i)}\log\frac{\exp{({z}_{i}\boldsymbol{\cdot}{z}_{p}/\eta)}}{\sum\limits_{b\in B(i)}\exp{({z}_{i}\boldsymbol{\cdot}{z}_{b}/\eta)}},
\end{split}
\end{equation}
where $z_{*}=\|Proj(z)\|_2$ means that the representation space is mapped to the unit hypersphere through multi-layer perceptron and ridge regularization, so that the similarity can be measured by the inner product. $\eta \in \mathcal{R}^{+}$ means a scalar temperature parameter. $i$ is the index of anchor. $B(i)\equiv \{1,\dots,2N\}\setminus\{i\}$ is the set of indices of all samples among the 2N trajectories, excluding the index $i$. $P(i)\equiv\{p\in B(i): {Label}(p)={Label}(i)\}$ is the set of indices of all positive samples among the $2N$ trajectories.

By completing contrastive learning accompanied by supervision signals, we effectively align a more discriminative representation space, where the trajectory representations of the same user are closer, and those of different users are further apart. Subsequently, we discard the projection layer, freeze the parameters of the embedding layer and dual-stream encoder $\mathcal{F}_{\theta^{\blacktriangle}_{\alpha}}$ and $\mathcal{F}_{\theta^{\blacktriangle}_{\beta}}$, and input all the trajectories of each user in the training set into the backbone network to obtain the individual trajectory representations:
\begin{equation}
{z^{\blacktriangle}}=\|(\gamma \cdot {\mathcal{F}(\theta^{\blacktriangle}_{\alpha};~\mathbf{X})}+(1-\gamma) \cdot {\mathcal{F}(\theta^{\blacktriangle}_{\beta};~\mathbf{X})})\|_2.
\end{equation}

Considering that traditional supervised loss typically uses a fixed cross-entropy loss classifier, it does not scale well to dynamic and large-scale trajectory-user linking. Therefore, we propose using a cosine similarity loss to avoid this issue. Specifically, by aggregating the trajectory representations of a single user through a pooling layer, we can precompute the user labels needed for the next stage of training:
\begin{equation}
{u}_{i}=\operatorname{Average-Pooling}(\{z^j|j\in U_i\}),
\end{equation}
where $z^j$ means all trajectory representations of user $U_i$.

\subsection{Trajectory-User Linking Layer}
To link unknown trajectories to the users who generated them, we further designed a TUL (Trajectory-User Link) predictor, composed of activation functions and a multilayer perceptron, to obtain the final trajectory representation 
$y=f(\theta;~z)$. This projection layer is intended to further enhance the representational capacity of the input trajectories.

In this representation space, we minimize the cosine similarity loss between the representation of the input trajectory and the corresponding user representation: 
\begin{equation}
 \begin{split}
     \mathcal{L}_{ul} & =\sum_{i\in N}(1-\cos(y_{i},u_{i})),
     \end{split}
\end{equation}
where $u_{i}$ denotes the user label who generated trajectory $i$, $y_i$ means unlinked trajectory representation, $N$ is number of the trajectories. During the testing stage, by calculating the similarity between the input trajectories and the corresponding user representations, each input trajectory is linked with the most similar user label.

\subsection{Model Training}

Our model uses a two-stage training pipeline to capture users' mobility characteristics and learn generalized features. By integrating semantic information from both raw and augmented trajectories, we obtain enriched user and trajectory representations.

In the first stage, the input trajectory is augmented, and both the input and augmented trajectories are passed through the dual-stream encoder separately to obtain their respective latent features. These features are then projected into a lower-dimensional space via a projection layer, where they are normalized and used to compute the contrastive learning loss. In the second stage, we train trajectory-user linking layer by minimizing the cosine similarity loss between the representations of the trajectory and the corresponding user.

%% file: 5Experiments.tex
\section{Experiments}

\subsection{Datasets}

We use real-world check-in data collected from the popular location-based social network platform Foursquare~\cite{yang2014modeling}, selecting data from three cities and the entire United States for our dataset. For the cities of Tokyo (TKY), New York City (NYC), and Santiago, we select the top 800 and 400 users, respectively, who have the most trajectories, to evaluate model performance. To demonstrate the effectiveness of our proposed framework on a large scale, we select the top 5000 and 10000 users with the most trajectories across the United States for evaluation. In our experiments, we use the first 80\% of each user's sub-trajectories for training, and the remaining 20\% for testing. Additionally, 20\% of the training data is set aside as a validation set to assist with an early stopping mechanism to find the best parameters and avoid overfitting.
The statistics of all datasets are summarized in Table~\ref{tab:dataset_table}.

\input{  table/dataset}
\input{  table/performance}
\input{  table/US_performance}

\subsection{Approach for Comparison}
We compare \model with recent state-of-the-art models for trajectory user-linking. Specifically, the following baseline approaches are evaluated.
\begin{itemize}
    \item \textbf{KNNTUL}~\cite{najjar2022trajectory} -  is the first work to attempt solving the large-scale TUL problem by employing the k-Nearest Neighbors (KNN) method.
    \item \textbf{MainTUL}~\cite{chen2022mutual} - is a knowledge distillation framework to capture rich contextual features in check-in data, with an RNN encoder and a temporal-aware Transformer encoder.
    \item \textbf{S$^2$TUL}~\cite{deng2023s2tul} - is a semi-supervised framework that models the relationships between trajectories by constructing both homogeneous and heterogeneous graphs. The framework includes four variants, of which we select the two that performed best across most datasets: S$^2$TUL-R and S$^2$TUL-HRSTS.
\end{itemize}
\subsection{Evaluation Metrics and Parameter Settings}
We use the Acc@k, Macro-F1 and Micro-F1 to evaluate the model performance. Specifically, Acc@k is used to evaluate the accuracy of TUL prediction. Unlike previous work that used Macro-Precision and Macro-Recall, we opted to include Micro-F1 for the following reasons: Macro-F1 is derived from Macro-Precision and Macro-Recall, effectively integrating their results. Additionally, Micro-F1 offers an overall performance metric that more accurately reflects the model's behavior across the entire dataset.
For the baselines, we use the parameter settings recommended in their papers and fine-tune them. For \model, we set default embedding dimension to 512, apply an early stopping mechanism with patience to 5 to avoid over fitting, and adjust the learning rates as follows: In the first stage, the initial learning rate is set to 0.001 and decays by 20\% every 5 epochs. In the second stage, the initial learning rate is set to 0.0005 and decays by 90\% every 5 epochs.The source code of our model is available at \url{https://github.com/sleevefishcode/ScalableTUL}.

All experiments  are conducted  on  a  machine  with  Intel(R) Xeon(R) Silver 4214 (2.20GHz 12 cores) and NVIDIA GeForce RTX 3090 (24GB Memory).
\subsection{Overall Performance}
We report the overall performance with baselines in Table~\ref{tab:deep_perform_compared} and Table~\ref{tab:US_perform_compared}, where the best is shown in bold and the second best is shown as underlined. 

As shown in Table~\ref{tab:deep_perform_compared}, \model significantly
outperforms all baselines in terms of all evaluation metrics on the check-in datasets of three cities, except for Acc@5 on TKY. This performance can be attributed to the spatio-temporal augmentation strategy and the dual-stream trajectory encoder. The former helps to alleviate issues related to trajectory irregularity and sparsity, thereby enhancing the model's generalization capability, while the latter more effectively captures the spatio-temporal movement patterns in users’ check-in trajectories, accommodating different trajectory lengths.
% \link{Code}{https://github.com/sleevefishcode/ScalableTUL}
% % \link{Datasets}{https://aaai.org/example/datasets}
% % \link{Extended version}{https://aaai.org/example/extended-version}

As the number of users in a dataset increases, the challenge of correctly linking users to their trajectories intensifies, leading to performance degradation in all models. However, \model can still achieve the best results. Specifically, \model
achieves an average improvement of 6.68\% in Acc@1 and 5.2\% in Macro-F1 compared to the best-performing baseline on the TKY and NYC datasets when the user count reaches 800. The significant improvements in accuracy and F1 scores suggest that \model’s architecture, especially its spatio-temporal trajectory augmentation strategy and dual-stream encoder, effectively captures complex user mobility patterns even as the number of users scales up, demonstrating its robustness against the increased complexity of larger datasets.

The experimental results on the nationwide U.S. dataset are presented in Table~\ref{tab:US_perform_compared}.
Notably, \model continues to perform well even when the number of users reaches 10k. In contrast, S$^2$TUL faces challenges in managing the large volume of user-generated trajectories, and the performance of the KNNTUL method falls significantly short of our model’s performance. For 5k users, our model surpasses the previously best-performing MainTUL model, with improvements of 6.89\% in Acc@1 and 6.11\% in Macro-F1. 

The consistent outperformance of \model across different user scales not only underscores its robustness but also demonstrates its scalability. The model's ability to maintain and even improve its performance as the dataset grows larger suggests that its architecture is well-suited for handling the increased complexity and data irregularities that come with larger user bases. These significant improvements emphasize the superiority of our approach in effectively learning and generalizing from diverse and extensive trajectory data, making \model a reliable and powerful solution for large-scale TUL applications.

\subsection{Ablation Study}

In our ablation study, we evaluate our model against five carefully designed variations:
(1) \textbf{w/o T-Aug} -- This variant excludes temporal augmentation from the trajectory augmentation strategy. (2) \textbf{w/o S-Aug} -- This variation omits the random mask used in trajectory augmentation. (3) \textbf{w/o Aug} -- Here, the entire trajectory augmentation strategy is removed, meaning no aggregation or masking is applied. (4) \textbf{w/o LT-Encoder} -- It removes the long-term encoder in dual-stream trajectory encoder.
(5) \textbf{w/o ST-Encoder} -- It excludes the short-term encoder in dual-stream trajectory encoder.

\begin{figure}
\centering

\includegraphics[width=0.99\columnwidth]{  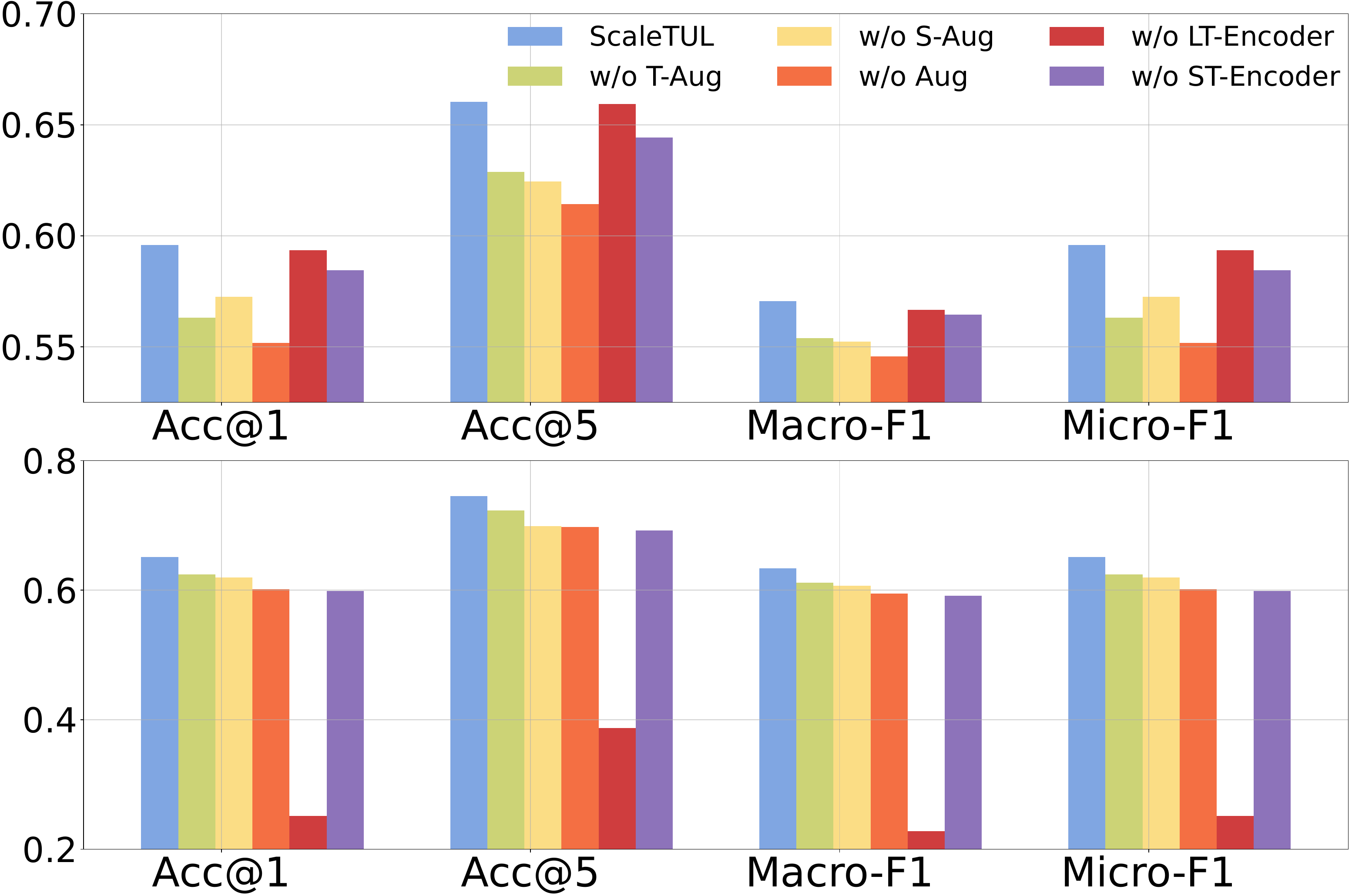}\\

\caption{Experimental results of component ablation on NYC (top) and TKY (bottom) ($|\mathcal{U}|=400$).}
\label{fig-ablation}
% \vspace{-3mm}
\end{figure}

The results of the ablation study are shown in Figure~\ref{fig-ablation}, which demonstrates the contribution of each component to the overall performance of our model. Notably, the long-term encoder has a significant affect on model performance, especially on the TKY dataset. This is primarily due to the fact that the TKY dataset includes a substantially higher number of longer trajectories (those with five or more points) compared to the NYC dataset. This highlights the effectiveness of our designed trajectory encoder, which includes the long-term component.
The impact of the trajectory augmentation strategy—\textbf{w/o T-Aug}, \textbf{w/o S-Aug}, and \textbf{w/o Aug}—underscores the importance. 
These variations reveal that temporal and spatial trajectory augmentations are crucial for improving model performance, particularly in handling data irregularities and sparsity. This is particularly evident in the %sparser%
NYC dataset, where trajectory augmentation helps mitigate issues related to data sparsity and irregularity, demonstrating its effectiveness in enhancing the model’s ability to generalize from limited data.

\subsection{Parameter Study}
\begin{figure}[b]
\centering

\begin{minipage}[b]{0.23\textwidth}
    \centering
    \includegraphics[width=0.99\textwidth]{  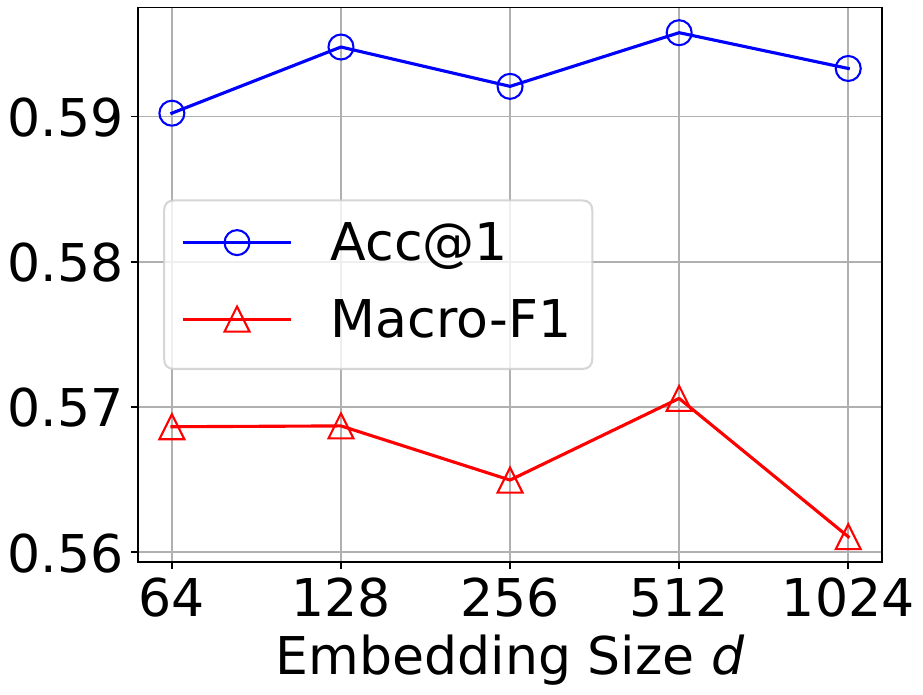}

    \label{fig-NYC-embed}
\end{minipage}
\hspace{0\textwidth} % Adjust space between the images
\begin{minipage}[b]{0.23\textwidth}
    \centering
    \includegraphics[width=0.99\textwidth]{  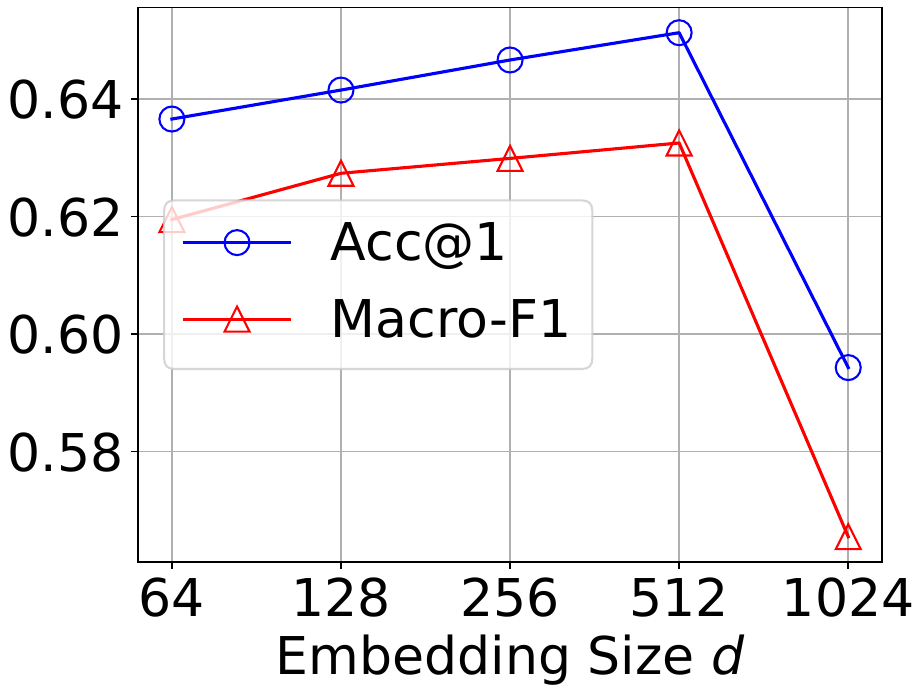}

    \label{fig-TKY-embed}
\end{minipage}

\caption{Result of parameter sensitivity embedding size $d$ on NYC(left) and TKY(right) ($|\mathcal{U}|=400$).}
\label{fig-parameter-embed}

\end{figure}

\begin{figure}[htbp]
\centering
\begin{minipage}[b]{0.23\textwidth}
    \centering
    \includegraphics[width=1\textwidth]{  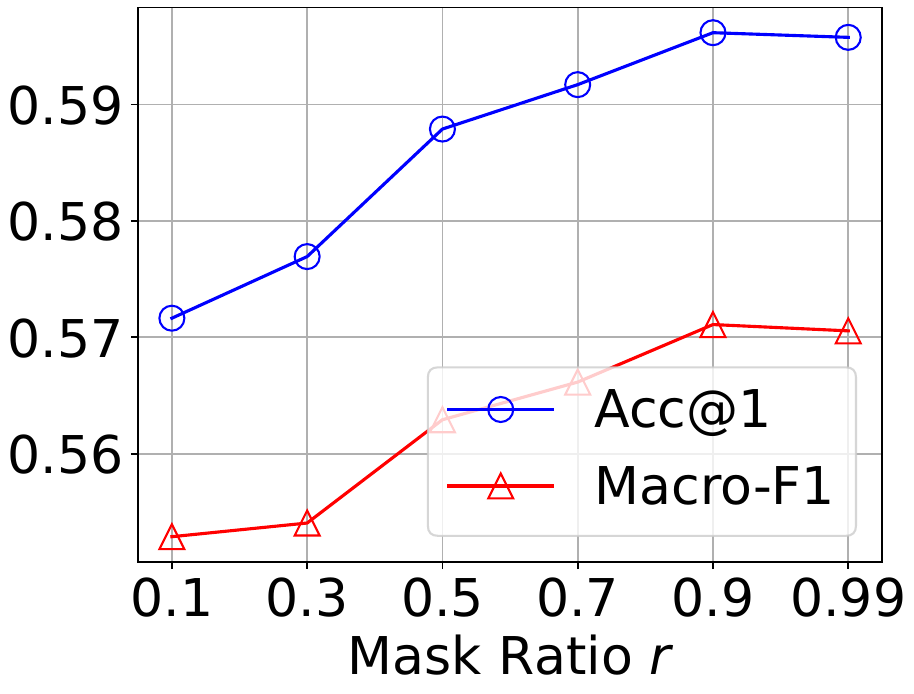}
    \vspace{-2mm}

    \label{fig-NYC-mask}
\end{minipage}
\hspace{0\textwidth} % Adjust space between the images
\begin{minipage}[b]{0.23\textwidth}
    \centering
    \includegraphics[width=1\textwidth]{  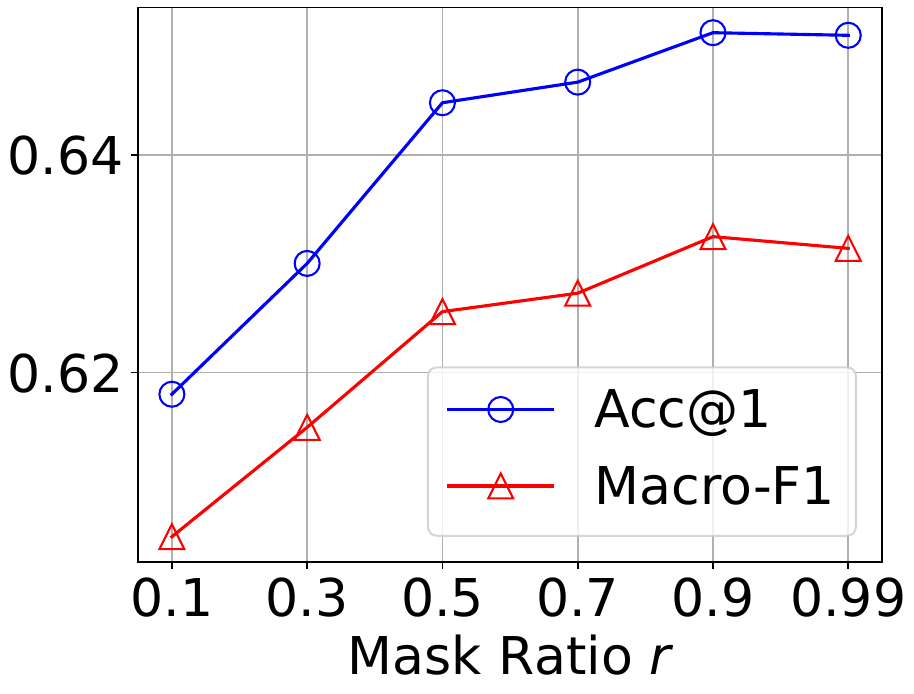}
    \vspace{-2mm}

    \label{fig-TKY-mask}
\end{minipage}

\caption{Result of parameter sensitivity mask ratio $r$ on NYC(left) and TKY(right) ($|\mathcal{U}|=400$).}
\label{fig-parameter-mask}
% \vspace{-2mm}
\end{figure}

We evaluate the model's sensitivity to embedding dimension $d$ and masking rate $r$. As shown in Figure~\ref{fig-parameter-embed}, the model's performance initially improves with $d$ increases and then declines. This trend occurs because a small embedding dimension is insufficient to capture the spatio-temporal dependencies intrinsic to trajectories, while an overly large embedding dimension reduces the model’s generalization capability. For masking rate $r$, larger values improve performance by alleviating data sparsity and boosting the model’s ability to generalize. However, when $r$ approaches 1, performance plateaus and eventually declines, as excessive masking eliminates important information that the model needs to make accurate predictions.
\subsection{Efficiency Study}
\begin{figure}

\begin{minipage}[b]{0.233\textwidth}
    \centering
    \includegraphics[width=\textwidth]{  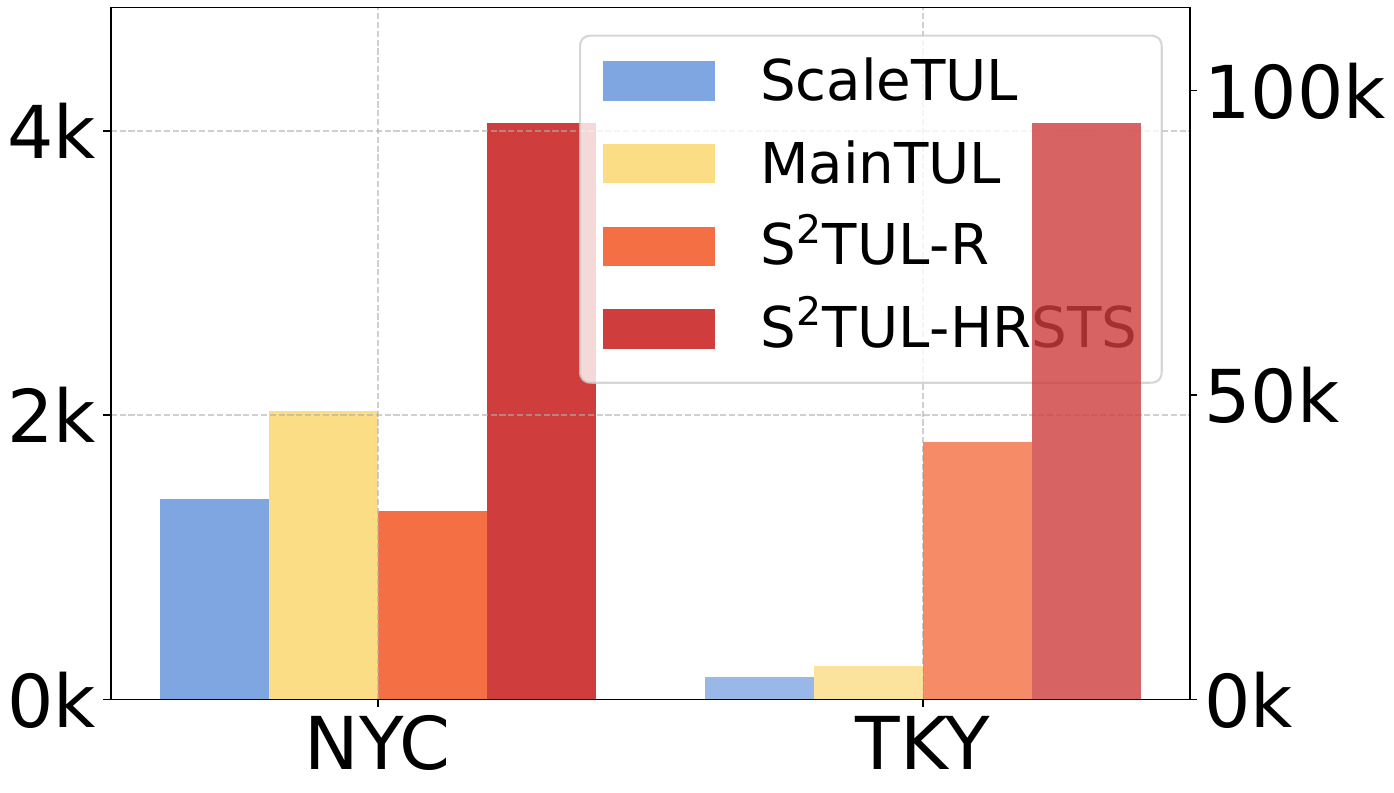}
    \vspace{-2mm}
 
    \label{time_study_400}
\end{minipage}
\hspace{-0.01\textwidth} % Adjust space between the images
\begin{minipage}[b]{0.233\textwidth}
    \centering
    \includegraphics[width=\textwidth]{  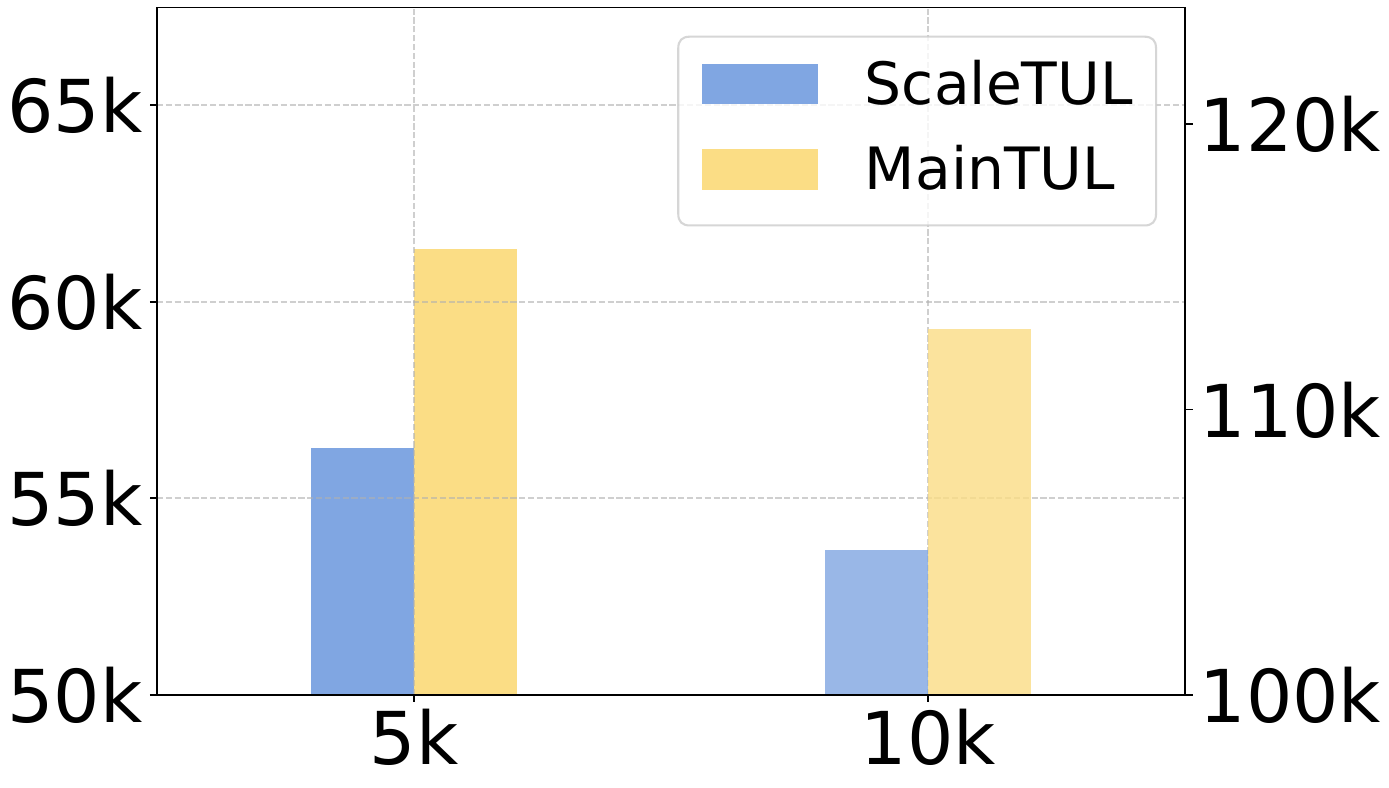}
    \vspace{-2mm}
    
    \label{time_study_US}
\end{minipage}
\vspace{-2mm}
\caption{Experimental results of running time, $|\mathcal{U}|=400$ (left) and $|\mathcal{U}|=5000/10000$ (right). }
\label{fig-time-study}
% \vspace{-3mm}
\end{figure}
We also perform efficiency experiments by comparing our model with the baseline models. As shown in Figure~\ref{fig-time-study}, the results indicate that our model outperforms other deep learning methods in terms of time efficiency. Particularly on the TKY dataset, our model achieve a speed improvement of approximately 96.04\% compared to the most effective baseline, S$^2$TUL-HRSTS. The reason for its significant time consumption is that constructing both homogeneous and heterogeneous graphs to integrate trajectory-level information becomes highly time-consuming when dealing with a large number of trajectories. 
It should be noted that when the number of users is 10k, the number of trainable parameters is reduced by 10 million compared to MainTUL, thereby lowering hardware demands and shortening training time. 

%% file: table/dataset.tex
\begin{table}[t!]
    % \vspace{-1mm}
    \begin{center}
    % \vspace{-2mm}
    \fontsize{9}{12} \selectfont
    \setlength{\tabcolsep}{0.9mm}{}	
    \begin{tabular}{c|c|c|c|c|c}
        \toprule
        \textbf{Datasets} & \#users  & \#trajectories & \#POIs & \#categories & Duration \\
        \midrule
        \multirow{2}{*}{TKY} & 800 & 104,413 & 39,698 & 239 & 11 months\\
        \cline{2-6}
        & 400 & 51,969 & 24,526 & 231 & 11 months\\
        \midrule
        \multirow{2}{*}{NYC} & 800 & 152,583 & 31,925 & 251 & 10 months\\
        \cline{2-6}
        & 400 & 39,784 & 19,903 & 248 & 10 months\\
        \midrule
        \multirow{2}{*}{Santiago} & 800 & 132,238 & 21,967 & 381 &18 months\\
        \cline{2-6}
        & 400 & 64,932 & 13,703 & 357 & 18 months\\
        \midrule
        \multirow{2}{*}{US} & 5000 & 749,443 & 258,865 & 427 & 18 months\\
         \cline{2-6}
        & 10000 & 1,167,786 & 344,274 & 428 & 18 months\\
  
        \bottomrule
    \end{tabular}
    % \vspace{-6mm}
    \caption{Statistics of the datasets.}
    % \vspace{-3mm}
    \label{tab:dataset_table}
    \end{center}
\end{table}

%% file: table/performance.tex
\begin{table*}[t]
% \vspace{-2mm}
\centering
\footnotesize
\fontsize{9}{12} \selectfont
\setlength{\tabcolsep}{2.5mm}{}	
\begin{tabular}{c|c|c|c|c|c|c|c|c|c}
\toprule

\multirow{2}{*}{Dataset} & \multirow{2}{*}{Methods} & Acc@1 & Acc@5  & Macro-F1& Micro-F1 & Acc@1 & Acc@5 & Macro-F1 & Micro-F1\\

\cline{3-10}

& & \multicolumn{4}{c|}{$|\mathcal{U}|=400$} & \multicolumn{4}{c}{$|\mathcal{U}|=800$}\\

\midrule
% TKY city
\multirow{5}{*}{TKY} &KNNTUL &44.32\%  &57.13\%  &43.76\%  &44.32\%  &37.45\%  &50.02\%  &36.21\%    &37.46\%\\

\cline{2-10}

&MainTUL &60.69\%  &69.80\%   &59.18\%  &60.69\%  &\underline{57.02}\%  &\underline{67.82}\%  &\underline{55.48}\%  &\underline{57.02}\%\\

\cline{2-10}

&$\rm{S^2TUL}$-$\rm{R}$ &59.98\%  &73.04\%  &60.82\%  &59.98\%  &OOM &OOM  &OOM &OOM\\

\cline{2-10}
&$\rm{S^2TUL}$-$\rm{HRSTS}$ &\underline{63.53}\%  &\textbf{75.90}\% &\underline{62.94}\%   &\underline{63.71}\%  &OOM  &OOM  &OOM  &OOM \\

\cline{2-10}
& \textbf{\model} &\textbf{65.13}\%$^*$  &\underline{74.47}\%$^*$  &\textbf{63.37}\%$^*$  &\textbf{65.13}\%$^*$&\textbf{60.63}\%$^*$  &\textbf{71.03}\% $^*$  &\textbf{58.74}\%$^*$  &\textbf{60.63}\%$^*$  \\

\midrule
% NYC city
\multirow{5}{*}{NYC} &KNNTUL &44.69\%    &54.31\%  &43.81\%  &44.68\%  &39.15\%  &49.95\%  &37.70\%  &39.15\%\\

\cline{2-10}

&MainTUL &54.84\%  &60.83\%  &53.36\%  &54.83\%  &\underline{52.43}\%  &59.50\%  &\underline{51.00}\%  &\underline{52.43}\%\\

\cline{2-10}

&$\rm{S^2TUL}$-$\rm{R}$ &\underline{57.00}\%  &64.35\%  &\underline{56.52}\%  &\underline{57.00}\%  &51.97\%  &\underline{60.52}\%  &50.56\%  &51.96\% \\
\cline{2-10}

&$\rm{S^2TUL}$-$\rm{HRSTS}$ &56.88\%  &\underline{65.52}\%  &54.65\%  &56.88\%  &51.59\%  &60.39\%  &49.00\%  &51.58\% \\

\cline{2-10}

&\textbf{\model} &\textbf{59.43}\% $^*$ &\textbf{66.03}\%$^*$  &\textbf{57.06}\%$^*$  &\textbf{59.43}\%$^*$  &\textbf{56.11}\%$^*$    &\textbf{63.78}\% $^*$ &\textbf{53.30}\%$^*$  &\textbf{56.11}\%$^*$\\

% Santiago city
\midrule
\multirow{5}{*}{Santiago} &KNNTUL &52.35\%    &62.78\%  &53.39\%  &52.34\%  &46.56\%  &57.41\%  &47.74\%  &46.57\%\\

\cline{2-10}

&MainTUL &\underline{55.29}\%  &64.59\%  &\underline{55.44}\%  &\underline{55.29}\%  &\underline{51.82}\%  &\underline{63.00}\%  &\underline{51.61}\%  &\underline{51.82}\%\\

\cline{2-10}

&$\rm{S^2TUL}$-$\rm{R}$ &53.96\%  &\underline{64.95}\%  &54.70\%  &53.96\%  &49.46\%  &61.40\%  &50.08\%  &49.46\% \\
\cline{2-10}

&$\rm{S^2TUL}$-$\rm{HRSTS}$ &54.00\%  &64.10\%  &53.34\%  &54.00\%  &OOM  &OOM  &OOM  &OOM \\

\cline{2-10}
&\textbf{\model} &\textbf{58.44}\%$^*$  &\textbf{67.12}\%$^*$  &\textbf{57.46}\%$^*$  &\textbf{58.43}\%$^*$  &\textbf{53.55}\%$^*$   &\textbf{64.51}\%$^*$  &\textbf{52.61}\%$^*$  &\textbf{53.55}\%$^*$\\

\bottomrule

\end{tabular}
\caption{Performance comparison with baselines on the dataset from three cities. OOM stands for ``out of memory". Marker * indicates the results are statistically significant (t-test with p-value $<$ 0.01).}
\label{tab:deep_perform_compared}
% \vspace{-1mm}
\end{table*}

%% file: table/US_performance.tex
\begin{table*}[t]
% \vspace{-1mm}
\centering
\footnotesize

% \vspace{-1mm}

\fontsize{9}{12} \selectfont
\setlength{\tabcolsep}{2.5mm}{}	
\begin{tabular}{c|c|c|c|c|c|c|c|c|c}
\toprule

\multirow{2}{*}{Dataset} & \multirow{2}{*}{Methods} & Acc@1 & Acc@5  & Macro-F1& Micro-F1 & Acc@1 & Acc@5 & Macro-F1 & Micro-F1\\

\cline{3-10}

& & \multicolumn{4}{c|}{$|\mathcal{U}|=5000$} & \multicolumn{4}{c}{$|\mathcal{U}|=10000$}\\
\midrule
% TKY city
\multirow{5}{*}{US} &KNNTUL &45.59\%  &57.36\%  &44.85\%  &45.59\% &39.00\%  &\underline{51.13}\%  &37.70\%  &39.01\%\\

\cline{2-10}

&MainTUL  &\underline{51.40}\%  &\underline{57.71}\%  &\underline{50.89}\%  &\underline{51.39}\% &\underline{43.85}\%  &51.01\%  &\underline{42.83}\%  &\underline{43.85}\%\\

\cline{2-10}

&$\rm{S^2TUL}$-$\rm{R}$   &OOM &OOM  &OOM &OOM&OOM&OOM&OOM&OOM\\

\cline{2-10}
&$\rm{S^2TUL}$-$\rm{HRSTS}$   &OOM  &OOM  &OOM  &OOM&OOM&OOM&OOM&OOM \\
\cline{2-10}
& \textbf{\model}&\textbf{54.94}\%$^*$  &\textbf{64.05}\% $^*$  &\textbf{54.02}\%$^*$  &\textbf{54.94}\%$^*$  &\textbf{44.90}\%$^*$ &\textbf{53.88}\%$^*$ &\textbf{44.32}\%$^*$ &\textbf{44.90}\%$^*$\\

\bottomrule

\end{tabular}
\caption{Performance comparison with baselines on the dataset from the United
States. OOM stands for ``out of memory". Marker * indicates the results are statistically significant (t-test with p-value $<$ 0.01).}
\label{tab:US_perform_compared}
% \vspace{-4mm}
\end{table*}

%% file: 6Conclusion.tex
\section{Conclusion}

In this paper, we propose a novel scalable trajectory-user linking with dual-stream representation networks called ScaleTUL.
Experimental results on datasets from three cities and across the United States demonstrate that ScaleTUL outperforms the state-of-the-art methods and proves its efficiency in efficiency study.